\documentclass[10pt,twocolumn,letterpaper]{article}

\usepackage{cvpr}
\usepackage{times}
\usepackage{epsfig}
\usepackage{graphicx}
\usepackage{amsmath}
\usepackage{amssymb}
\usepackage{epsf, amsmath, multirow}

\usepackage[breaklinks=true,bookmarks=false]{hyperref}

\cvprfinalcopy % *** Uncomment this line for the final submission

\def\cvprPaperID{****} % *** Enter the CVPR Paper ID here

\ifcvprfinal\fi
\begin{document}

%%%%%%%%% TITLE
\title{Embedding Human Knowledge into Deep Neural Network via Attention Map}

\author{
    Masahiro Mitsuhara$^\dagger$, Hiroshi Fukui$^\dagger$, Yusuke Sakashita$^\dagger$,\\
    Takanori Ogata$^\ddagger$, Tsubasa Hirakawa$^\dagger$, Takayoshi Yamashita$^\dagger$, Hironobu Fujiyoshi$^\dagger$\\
    $^\dagger$Chubu University, $^\ddagger$ABEJA Inc.\\
    {\tt \small \{mitsuhara@mprg.cs, fhiro@mprg.cs, sakashita@mprg.cs\}.chubu.ac.jp,}\\
    {\tt \small takanori@abejainc.com,}\\
    {\tt \small \{hirakawa@mprg.cs, takayoshi@isc, fujiyoshi@isc\}.chubu.ac.jp}\\
}

\maketitle
%\thispagestyle{empty}

%%%%%%%%% ABSTRACT
\begin{abstract}
% In this work, we aim to realize a method for embedding human knowledge in deep neural networks. While the conventional method to embed human knowledge has been applied for non-deep machine learning, it is difficult to apply it for deep learning models due to the enormous number of model parameters. To tackle this problem, we focus on the attention mechanism of an attention branch network (ABN). In this paper, we propose a fine-tuning method that utilizes an attention map which is manually edited by human expert. Our fine-tuning method can train a network so that the output attention maps correspond to the edited ones. As a result, the fine-tuned network can output an attention map that takes into account human knowledge. Experimental results with ImageNet, CUB-200-2010, and IDRiD demonstrate that it is possible to obtain a clear attention map as a visual explanation from the judgment grounds and improve the classification performance. Our findings can be applied to a novel framework for optimizing networks through human intuitive editing via a visual interface and suggest new possibilities for human-machine cooperation in addition to the improvement of visual explanations.

In this work, we aim to realize a method for embedding human knowledge into deep neural networks.
While the conventional method to embed human knowledge has been applied for non-deep machine learning, it is challenging to apply it for deep learning models due to the enormous number of model parameters.
To tackle this problem, we focus on the attention mechanism of an attention branch network (ABN).
In this paper, we propose a fine-tuning method that utilizes a single-channel attention map which is manually edited by a human expert.
Our fine-tuning method can train a network so that the output attention map corresponds to the edited ones.
As a result, the fine-tuned network can output an attention map that takes into account human knowledge.
Experimental results with ImageNet, CUB-200-2010, and IDRiD demonstrate that it is possible to obtain a clear attention map for a visual explanation and improve the classification performance.
Our findings can be a novel framework for optimizing networks through human intuitive editing via a visual interface and suggest new possibilities for human-machine cooperation in addition to the improvement of visual explanations.

\end{abstract}

%%%%%%%%% BODY TEXT
%%%%%%%%%%%%%%%%%%%%%%%%%%%%%%%%%%%%%%%%%%%%%%%%%%%%%%%%%%%%%%%%%%%%%%%%
% Introduction
%%%%%%%%%%%%%%%%%%%%%%%%%%%%%%%%%%%%%%%%%%%%%%%%%%%%%%%%%%%%%%%%%%%%%%%%
\section{Introduction}
% 視覚的説明
% 深層学習の判断根拠を解析する方法として，視覚的説明~\cite{ribeiro2016should,Aditya2017,Selvaraju2017,Smilkov2017,Zeiler2014,Zhou2016,fukui2018attention,montavon2018methods,Jost2015,Petsiuk2018rise}がある．
% 視覚的説明は，Convolutional Neural Network~(CNN)~\cite{LeCun1989,Alex2014}の推論時における判断根拠を，注視領域をヒートマップで表現したAttention mapにより解析する．
% 視覚的説明の代表的な手法として，畳み込み層の応答値を用いてAttention mapを獲得するClass Activation Mapping~(CAM)\cite{Zhou2016}や，Attention map獲得時に特定クラスの正値の勾配を重み値として用いるGradient weighted-CAM (Grad-CAM)~\cite{Selvaraju2017}，視覚的説明のAttention map をAttention機構へ応用するAttention Branch Network~(ABN)~\cite{fukui2018attention}が提案されている．
% これらの視覚的説明の手法により，CNNの判断根拠の解釈が可能となりつつある．

\begin{figure}[tb]
\centering
\includegraphics[width=0.95\linewidth]{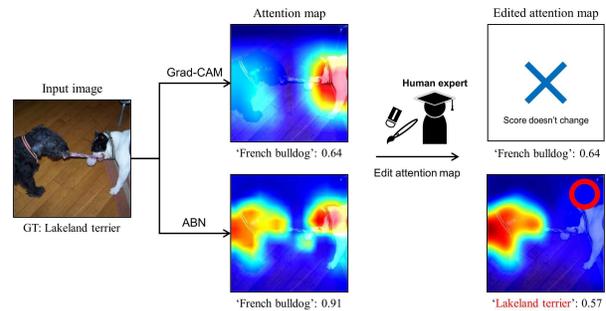}
% \caption{Attention mapの修正によるスコアの変化．}
\caption{Adjustment of recognition result by editing an attention map on visual explanation.}
\label{fig:topview_attention_map}
\end{figure}

Visual explanation is often used to interpret the decision-making of deep learning in the computer vision field \cite{ribeiro2016should,chattopadhay2018grad,Selvaraju2017,smilkov2017smoothgrad,Zeiler2014,Zhou2016,fukui2018attention,montavon2018methods,Jost2015,fong2017interpretable,Petsiuk2018rise,wagner2019interpretable}. Visual explanation analyzes the decision-making of a convolutional neural network (CNN) \cite{LeCun1989, Alex2014} by visualizing an attention map that highlights discriminative regions used for image classification. 
% todo: 細かいので，ざっくりと説明しても良いかも
Typical visual explanation approaches include class activation mapping (CAM) \cite{Zhou2016} and gradient weighted-CAM (Grad-CAM) \cite{Selvaraju2017}. CAM outputs an attention map by utilizing the response of the convolution layer. Grad-CAM outputs an attention map by utilizing the positive gradients of a specific category. Attention branch network (ABN) \cite{fukui2018attention} that extends an attention map to an attention mechanism in order to improve the classification performance has also been proposed.
% As typical visual explanations, class activation mapping~(CAM)~\cite{Zhou2016}, which outputs an attention map by using the response of the convolution layer, gradient weighted-CAM~(Grad-CAM)~\cite{Selvaraju2017}, which outputs an attention map by using the positive gradients of a specific category, and the Attention Branch Network~(ABN)~\cite{fukui2018attention}, which extends an attention map to an attention mechanism, have been proposed.
Thanks to these visual explanation methods, the decision-making of CNNs is becoming clearer.
%
% しかし，Ground truth~(GT)とこの視覚的説明である注視領域が一致しない場合が発生する．
% 図~\ref{fig:topview_attention_map}の例は，ImageNet DatasetにおけるGrad-CAMとABNのAttention mapの例である．
% この例では，画像にGTとして``Lakeland terrier"のラベルが付与されているが，画像中には``Lakeland terrier"と``French bulldog"の2つの物体が存在する．
% そのため，GTと異なる物体に注視領域が発生すると誤認識を誘発する．
% 例えば，医療画像診断の場合，病気か否かを画像から判断する際に疾患領域を注視する．
% そのため，出力結果と注視領域の不一致は判断根拠を説明するのに深刻な問題となり，医療画像診断における信頼性の低下へと繋がる．
%
However, an inconsistency between the target region of the recognition result, namely the ground truth (GT), and an attention region may occur. Examples of attention maps generated by Grad-CAM and ABN are shown in Fig.~\ref{fig:topview_attention_map}. Although the input image is annotated ``Lakeland terrier" as a GT, it contains multiple objects: ``Lakeland terrier" and ``French bulldog". Therefore, if CNN pays attention to different objects than the GT, it is likely to perform incorrect classifications. This mismatch would be critical in some applications. For example, in medical image recognition systems, a mismatch between the classification result and the attention region would degrade the reliability of the classification.

% However, an inconsistency between the recognition result, namely grand truth (GT), and an attention region may occur.
% Examples of attention maps generated by Grad-CAM and ABN are shown in Fig.~\ref{fig:topview_attention_map}.
% Although the input image is annotated ``Lakeland terrier" as a GT, it contains multiple objects: ``Lakeland terrier" and ``French bulldog".
% Therefore, if the CNN pays attention to different objects than the GT, it is likely to perform incorrect classifications.
% This mismatch would be critical in some applications.
% For example, in medical image recognition systems, the mismatch between the classification result and attention region may degrade the reliability of the classification system.

% データの問題 & 視覚的説明
% この問題を解決するために，本研究ではHuman in the loop~(HITL)のフレームワークを用いる．
% HITLとは，機械学習に人の知見を取り入れることで，機械学習における認識が困難な課題を容易にするアプローチである．
% HITLのフレームワークを取り入れた手法は数多く提案されている~\cite{branson2010visual,deng2013fine,wang2016human,branson2011strong,donahue2011annotator,parkash2012attributes,parikh2011interactively,duan2012discovering,sorokin2008utility}．
% 従来のHITLを導入した手法は，決定木やConditional Random Fields (CRF)~\cite{Quattoni2007}等の比較的パラメータの少ない機械学習モデルで用いられてきた．
% しかしながら，様々な画像認識分野で用いられる深層学習を，HITLのフレームワークに導入することは非常に困難である．
% これは，深層学習が非常に膨大なパラメータを持つためである．
% HITLを深層学習へ応用するには，人の知見をこれらの膨大のパラメータに反映する必要がある．

To solve this issue, we aim to realize a method for embedding human knowledge into deep learning models. Although this approach has been widely proposed \cite{branson2010visual,deng2013fine,wang2016human,branson2011strong,donahue2011annotator,parkash2012attributes,parikh2011interactively,duan2012discovering,sorokin2008utility}, the conventional methods are based on rather small machine learning models featuring fewer model parameters, such as decision trees and conditional random fields (CRFs) \cite{Quattoni2007}. It is difficult to embed human knowledge into deep learning models due to the massive number of model parameters.

% todo: これはなくても良いかもしれない（上の文の裏返し）
% Therefore, to embed human knowledge to deep neural network, we need to reflect human knowledge to the massive parameters.

% To resolve this issue, we use a human-in-the-loop (HITL) framework.
% HITL is a machine learning approach that enables difficult image recognition tasks to be easily trained by introducing human knowledge~\cite{branson2010visual,deng2013fine,wang2016human,branson2011strong,donahue2011annotator,parkash2012attributes,parikh2011interactively,duan2012discovering,sorokin2008utility}.
% Conventional HITL approaches have been used in small machine learning models such as decision trees and conditional random fields~(CRFs)~\cite{Quattoni2007}, but it is difficult to use them in deep learning that is used in various computer vision tasks, because the deep learning models have massive parameters.
% To apply HITL to deep learning-based approaches, we need to reflect human knowledge to the massive parameters of deep learning.

% HITLのフレームワークにより人の知見を大規模な深層学習へ取り入れた手法として，Linsleyらの手法~\cite{linsley2018learning}が提案されている．Linsleyらの手法は，大規模なモデルであるSqueeze-and-Exitiation Network~(SENet)~\cite{Hu2017}のAttention機構に空間のAttention機構を追加し，Attention機構の重みが人の知見を取り入れたClickMe mapと同一になるように誤差を算出して学習させている．

% しかし，Linsleyらの手法ではAttention機構の重みに対して人の知見を取り入れるため，ダイレクトかつインタラクティブに人の知見を取り入れることが困難である．また，誤認識したサンプルにおいて，正しく認識するためにパラメータ等の調整による出力結果の調整が不可能である．

To this end, we focus on the visual explanation and the attention mechanism of ABN \cite{fukui2018attention}. ABN applies an attention map for visual explanation to the attention mechanism. Therefore, by editing an attention map manually, as shown in Fig.~\ref{fig:topview_attention_map}, ABN can output a desirable recognition result by inference processing using the edited attention map. We propose a fine-tuning method based on the characteristics of ABN and an edited attention map. The proposed method fine-tunes the attention and perception branches of ABN to output the same attention map as the edited one. By learning the edited attention map that incorporates human knowledge, we can both obtain a more interpretable attention map and improve the recognition performance.

% Our deep learning-based HITL approach calculates the training loss from the attention map output from ABN and the modified attention map, then we fine-tune the attention and perception branches of ABN by using the loss.
% By learning the attention map that incorporates human knowledge, it is expected to output highly explanatory attention map and improve the recognition performance.

%我々はこれを実現するために，視覚的説明とAttention機構を導入したABNの構造に着目する．
%ABNは，視覚的説明のAttention mapをAttention機構に応用している．
%そのため，ABNは図~\ref{fig:topview_attention_map}のようにAttention mapを人手によりエディットすることで，その修正したAttention mapに対する好ましい認識結果を得ることができる．
%そこで，本論文ではABNのこの特性を用いて，人手により修正したAttention mapを用いたABNの再学習法を提案する．
%提案する深層学習ベースのHITLでは，人手で修正したAttention mapとABNが出力したAttention mapから学習誤差を算出し，ABNの後層の2つのブランチをファインチューニングする．
%人の知見を取り入れたAttention mapを学習することで，より説明性の高いAttention mapの出力と認識性能の向上が期待できる．

Our contributions are as follows:
%以下に，本稿の貢献を示す．
\begin{itemize}
% \item We show that the recognition result is improved by manual modification the attention map of ABN which is a visual explanation, to reflect human knowledge.
\item We demonstrate that manually editing the attention map used for a visual explanation can improve the recognition performance by reflecting human knowledge.
%\item 視覚的説明であるABNのAttention mapを手動で修正することにより，人の知見を反映し，認識結果が改善することを示す．
%
% \item We propose a method for re-learning the network using an attention map that has been manually modified. we achieved introduction of human knowledge (HITL) in deep learning by learning that the attention map output by the network matches the modified attention map.
\item We propose a fine-tuning method that uses manually edited attention maps. By training a network to output the same attention maps as the edited ones, we can embed human knowledge into deep neural networks.
%\item 人手で修正したAttention mapを用いて，ネットワークを再学習する手法を提案する．ネットワークが出力するAttention mapが修正したAttention mapと一致するように学習することで，深層学習における人の知見の導入(HITL)を実現する．
%
% \item In this paper, we will improve the explanation that has been widely required in Deep neural networks in recent years. Furthermore, we formulate a framework for optimizing networks through human intuitive modification via a visual interface and show new possibilities for human-machine cooperation.
\item Beyond the visual explanation widely required in the development of deep neural networks, this paper formulates a novel optimization framework of networks that humans can intuitively edit via a visual interface. This will open new doors to future human-machine cooperation.
%\item Deep Neural Networkにおいて近年広く求められている説明性の実現に留まらず，視覚的インターフェースを介して人が直感的に修正することで，ネットワークを最適化するフレームワークを定式化し，人・機械協調の新たな可能性を示す．
\end{itemize}

%%%%%%%%%%%%%%%%%%%%%%%%%%%%%%%%%%%%%%%%%%%%%%%%%%%%%%%%%%%%%%%%%%%%%%%%
% Related work
%%%%%%%%%%%%%%%%%%%%%%%%%%%%%%%%%%%%%%%%%%%%%%%%%%%%%%%%%%%%%%%%%%%%%%%%
\section{Related work}
% \subsection{Human-in-the-loop on computer vision}
\subsection{Embedding human knowledge}

% 機械学習手法に人の知見を導入するメジャーな手法として，human-in-the-loop (HITL) \cite{branson2010visual,deng2013fine,wang2016human,branson2011strong,donahue2011annotator,parkash2012attributes,parikh2011interactively,duan2012discovering,sorokin2008utility}がある．
% HITLは学習の過程で人が介入するアプローチである．（以下は同じ）
One of the major approaches to embedding human knowledge into machine learning models is human-in-the-loop (HITL) \cite{branson2010visual, deng2013fine, wang2016human, branson2011strong, donahue2011annotator, parkash2012attributes, parikh2011interactively, duan2012discovering, sorokin2008utility}. In HITL, human operators intervene during the training of machine learning models. In the field of computer vision, HITL is often applied to difficult recognition tasks such as fine-grained recognition. Several feature extraction approaches based on human knowledge have been proposed~\cite{branson2010visual, duan2012discovering, deng2013fine}.

% 人の知見を学習に導入するHITLな手法~\cite{branson2010visual,deng2013fine,wang2016human,branson2011strong,donahue2011annotator,parkash2012attributes,parikh2011interactively,duan2012discovering,sorokin2008utility}は，数多く提案されている．
% HITLとは，機械では認識が困難なタスクを効率的に学習するために，人の知見を機械学習に導入する方法である．
% 画像認識分野では詳細認識等の認識が困難なタスクにHITLが導入されており，特徴抽出を人の知見をヒントに獲得する手法が提案されている~\cite{branson2010visual,duan2012discovering,deng2013fine}．
% For training difficult recognition tasks, HITL that introduces human knowledge to machine learning has been widely studied~\cite{branson2010visual, deng2013fine, wang2016human, branson2011strong, donahue2011annotator, parkash2012attributes, parikh2011interactively, duan2012discovering, sorokin2008utility}.
% In the field of computer vision, HITL is often applied to difficult recognition tasks such as fine-grained recognition, and several feature extraction approaches based on human knowledge have been proposed~\cite{branson2010visual, duan2012discovering, deng2013fine}.

% 詳細認識におけるHITLでは，様々な人の知見が機械学習へと導入されている．
% Branson らは，特定の鳥に対する質問の回答を人の知見として使用し，決定木の学習を補助する対話型HITLの手法を提案している~\cite{branson2010visual}．
% ターゲットの属している項目を人の知見として扱うのではなく，ターゲットの特徴的な位置や領域を人の知見として扱う方法も提案されている．
% Duan ら~\cite{duan2012discovering}は，鳥の色と部位を人の知見として扱い，これらの情報をConditional Random Fields~(CRF)~\cite{Quattoni2007}の学習に導入する．
% Deng らは，2種類の鳥をユーザに見せ，見分ける上で着目した領域を円形のBounding box~(以下，babble)でアノテーションした情報を人の知見として扱う~\cite{deng2013fine}．
% 鳥のペアや様々なユーザからbabbleを獲得することで，鳥のカテゴリを認識する上で特徴的な領域を獲得し，HITLのフレームワークに導入している．
% これらの位置情報は，鳥のカテゴリを分類する上で重要な位置を機械学習の学習へ導入できるため，高い精度で詳細認識が可能になる．

In HITL for fine-grained recognition, various kinds of human knowledge are introduced. Branson {\it et al.}~\cite{branson2010visual} proposed an interactive HITL approach that helps to train a decision tree by using a question and answer with respect to a specific bird. In addition to items inherent in an object, characteristic positions or regions of an object have also been used as human knowledge. Duan {\it et al.}~\cite{duan2012discovering} introduced the body part position and color of a bird as human knowledge into the training of a CRF. Deng {\it et al.}~\cite{deng2013fine} used a bubble, that is, a circular bounding box, as human knowledge. This bubble information is annotated from an attention region when a user distinguishes the two types of birds. By annotating the bubble with various pairs and users, characteristic regions of bird images can be obtained when we recognize bird categories. These bubbles are introduced to the HITL framework as human knowledge, and can improve the accuracy of fine-grained recognition because the machine learning model is trained with an important location for recognizing the bird category.
%
%しかしながら，HITLは少数のパラメータを持つモデルに適用されることは多いが，深層学習に適用されることは少ない．
%これは，深層学習が膨大なパラメータを持つことが原因であり，人の知見を反映するパラメータが膨大であるため，調整が困難なためである．
%
However, these methods have primarily been applied to models having a small number of parameters, and are rarely applied to deep learning. This is because deep learning has an enormous number of parameters, and it is difficult to adjust to such a high number.

%HITLのフレームワークにより人の知見を大規模な深層学習へ取り入れた手法として，Linsleyらの手法~\cite{linsley2018learning}が提案されている．
%Linsleyらの手法は，Squeeze-and-Exitiation Network~(SENet)~\cite{Hu2017}のAttention機構~\cite{Thang2015,Xu2015,Hu2017,Bahdanau2016,he2016deep,Volodymyr2014,Wang2017a,vaswani2017attention,Wang2018b,Zichao2016,You2016}に空間のAttention機構を追加し，人の知見を取り入れたClickMe mapを用いて，ResNetの数カ所に導入された全Attention機構の重みがClickMe mapと同一になるように誤差を算出して学習させている．
%また，Linsleyらの手法におけるAttention機構はchannel-wiseな構造であるため，各特徴マップに対してAttention mapが出力される．
%そのため，人がAttention mapを主観的に見たときに修正することが困難である．

Linsley {\it et al.}~\cite{linsley2018learning} proposed a method that incorporates human knowledge into large-scale deep neural networks using the HITL framework.
This method added a spatial attention mechanism into the attention mechanism~\cite{Thang2015,Xu2015,Hu2017,Bahdanau2016,Volodymyr2014,Wang2017a,vaswani2017attention,Wang2018b,Zichao2016,You2016} of squeeze-and-excitation networks~(SENet)~\cite{Hu2017} and trained the network by using a ClickMe map that introduces human knowledge to the weights of the attention mechanism. This method can achieve higher accuracy because the network is trained while the attention mechanism weights located at multiple points become the same as the ClickMe map.
Because the attention mechanism in \cite{linsley2018learning} is a channel-wise structure, attention maps are output for each feature map.
It is difficult to edit an attention map when a human operator views the map subjectively.
Meanwhile, we use a single-channel attention map for embedding human knowledge into deep neural networks.
A human operator can understand the attention map intuitively and edit the map through a visual interface interactively.

% With our method, the system interactively edits the attention map obtained from the ABN while a human operator confirms the recognition result through a visual interface.
% The attention map of the ABN is a single-channel map that indicates the area at which the network focuses on.
% Because the attention map is easy to understand for humans perceptually, humans can easily edit the attention map. Therefore, we focus on the network. By using the edited attention map including human knowledge, the proposed method fine-tunes it and improves the network performance.

%一方で，我々の手法はABNから出力されたAttention mapを視覚的インターフェースを介して人が認識結果を確認しながらインタラクティブに修正し，ABNをファインチューニングする．そのため，Linsleyらの手法よりダイレクトかつインタラクティブに人の知見を取り入れることができる．
%また，ABNの視覚的説明のAttention mapは人が主観的に見たときに理解しやすい．

%視覚的説明は，ネットワークが注視した位置をマップで表現したAttention mapを用いることで，深層学習の判断根拠を解析する．
%そのため，ABNのAttention mapはネットワークが認識時に注視した領域を示しているため，人が主観的に修正できる．
%提案手法では，人手によりAttention mapを修正し，修正したAttention mapを用いて再学習する．
%これにより，深層学習の学習に人の知見を導入し，ネットワークの性能改善を実現する．

% Visual explanation enables us to interpret the decision-making of deep learning using an attention map that highlights the attention region.
% Therefore, the attention map of the ABN can be modified subjectively by humans, because it indicates the area that the network gazes at the time of recognition.
% Our approach fine-tunes the CNN by using the modified attention map along with human knowledge and improves the network performance.

\subsection{Visual explanation}\label{sec:visual_explanation}

% To interpret the deep learning in computer vision ~\cite{Alex2014,lin2013network,szegedy2015going,Simonyan2014,Gao2017,Xie2017,Zagoruyko2016,Hu2017,he2016}, visual explanation that visualizes the attention region in the inference process has been used~\cite{ribeiro2016should,Aditya2017,Selvaraju2017,Smilkov2017,Zeiler2014,Zhou2016,fukui2018attention,montavon2018methods,Jost2015,Petsiuk2018rise}.
% CVの文献はいらないのでは？
To interpret deep learning in computer vision, visual explanation that visualizes the discriminative region in the inference process has been used~\cite{ribeiro2016should,chattopadhay2018grad,Selvaraju2017,smilkov2017smoothgrad,Zeiler2014,Zhou2016,fukui2018attention,montavon2018methods,Jost2015,fong2017interpretable,Petsiuk2018rise,wagner2019interpretable}. Visual explanation can be categorized into two approaches: gradient-based, which outputs an attention map using gradients, and response-based, which outputs an attention map using the response of the convolutional layer. One of the gradient-based approaches is Grad-CAM~\cite{Selvaraju2017}, which can obtain an attention map for a specific category by using the response of the convolution layer and a positive gradient in the backpropagation process. Grad-CAM can be applied to various pre-trained models.

%深層学習における画像認識分野~\cite{Alex2014,lin2013network,szegedy2015going,Simonyan2014,Gao2017,Xie2017,Zagoruyko2016,Hu2017,He2016}では，ネットワークの注視領域をマップで表現したAttention mapにより深層学習の判断根拠を解析する視覚的説明\cite{ribeiro2016should,Aditya2017,Selvaraju2017,Smilkov2017,Zeiler2014,Zhou2016,fukui2018attention,montavon2018methods,Jost2015,Petsiuk2018rise}が提案されている．
%視覚的説明には，勾配を用いてAttention mapを獲得する手法と，ネットワークの応答値からAttention mapを獲得する手法がある．
%勾配ベースの視覚的説明として，Grad-CAM~\cite{Selvaraju2017}が提案されている．
%Grad-CAMは，特定のクラスにおける正値の勾配のみを用いることで，Attention mapを獲得する．
%Grad-CAMの特徴として，学習済みの様々なネットワークからAttentioin mapを獲得することができる．

One of the response-based approaches is CAM~\cite{Zhou2016}, which outputs an attention map by using a $K$~channel feature map from the convolution layer of each category. The attention maps of each category are calculated by using the $K$~channel feature map and the weight at a fully connected layer. However, CAM degrades the recognition accuracy because spatial information is removed due to the global average pooling (GAP)~\cite{DBLP:journals/corr/LinCY13} layer between the convolutional and fully connected layers. To address this issue, ABN has been proposed~\cite{fukui2018attention}, which extends an attention map for the visual explanation to an attention mechanism. By applying an attention map to the attention mechanism, ABN improves the classification performance and obtains an attention map simultaneously.

%ネットワークの応答値のみ用いる視覚的説明として，CAM~\cite{Zhou2016}が提案されている．
%CAMでは，畳み込み層から得られるクラス数枚$K$の特徴マップを用いることでAttention mapを獲得する．
%Attention mapを獲得する際は，$K$~チャンネルの畳み込み層で得られた特徴マップと全結合層の重みを用いることで，各クラスにおけるAttention mapを獲得する．
%しかし，CAMは畳み込み層と全結合層の間のGlobal Average Pooling (GAP)~\cite{lin2013network}により空間情報が削除されるため，画像認識精度が低下しやすい．
%これらのCAMの問題を解決した手法として，ABNが提案されている．
%ABNは，CAMをベースに出力された視覚的説明におけるAttention mapをAttention機構へ応用するネットワークである．Attention機構は，特徴量への重み付けにより特定の領域の特徴に着目させ，ネットワークの汎化性能を向上させる方法である．
%ABNは，Attention mapをAttention機構へ応用することで，視覚的説明におけるAttention mapの出力と，画像認識精度の改善を同時に実現できる．

% todo: 下の2段落をまとめる
% 本論文では，ABNの構造に着目する．ABNのAttention機構により，人手で修正したattention mapを考慮した認識結果の調整が可能である．さらに，修正したattention mapとABNが出力するattention mapが同一となるように再学習することで，深層学習における人の知見を導入する．
In this paper, we focus on this ABN ability. Because of the attention mechanism, ABN can adjust recognition results by considering the manually edited attention map. Moreover, we propose a method for embedding human knowledge into the network by fine-tuning so that the edited attention map and the attention map obtained from ABN become the same.

% Therefore, in cases where an attention map different from the desired map is output, these visual explanation methods cannot adjust the recognition score by modifying the output map.
% In contrast, ABN can adjust the recognition result because it applies an attention map to the attention mechanism.
% %
% In this paper, we focus on this ABN ability.
% Attention mechanism of ABN makes it possible to obtain recognition results considering the modified attention map.
% In addition, we achieve introduction of human knowledge in deep learning by re-learning so that the modified attention map and the attention map output by ABN are the same.

%加えて，ABNはCAMやGrad-CAMとは異なり，Attention mapの修正に対応した認識結果の獲得が可能である．
%一方で，ABNはAttention mapをAttention機構へ応用するため，Attention mapの修正に応じた認識結果の調整が可能である．
%
%本研究では，このABNの構造に着目する．
%ABNのAttention機構により，修正したAttention mapを考慮した認識結果の獲得が可能である．
%さらに，修正したAttention mapとABNが出力するAttention mapが同一となるように再学習することで深層学習における人の知見の導入を実現する．

%%%%%%%%%%%%%%%%%%%%%%%%%%%%%%%%%%%%%%%%%%%%%%%%%%%%%%%%%%%%%%%%%%%%%%%%
% Investigation of editing attention map
%%%%%%%%%%%%%%%%%%%%%%%%%%%%%%%%%%%%%%%%%%%%%%%%%%%%%%%%%%%%%%%%%%%%%%%%
\section{Investigation of editing attention map}\label{sec:edit_attention}
% 本章では，ABNが出力するAttention mapを人手により修正した際のネットワークの挙動の変化を調査する．
% ABNは，出力されたAttention mapをAttention機構へ応用するため，Attention mapの修正に対応して認識結果が変化する特性がある．
% この特性の有効性を確認するために，Attention mapを人の知見により手動で修正することで，より適切な認識結果を得られるかをImageNet Datasetを用いて調査する．

With ABN, we should be able to adjust the recognition result by editing an attention map, since the attention map is used for the attention mechanism (as mentioned above). In this section, we investigate the behavior of ABN in a case where we edit an attention map manually. Specifically, we confirm the changes in classification performance by editing an attention map on the ImageNet dataset \cite{Deng2009}.

% In this section, we investigate the behavior of ABN when we edit an attention map manually.
% Since an attention map is applied to an attention mechanism, ABN adjusts the recognition result by modifying the attention map.
% To confirm this characteristic, we investigate the changes of classification performance by modifying an attention map with the ImageNet dataset \cite{Deng2009}.

% そこで，予備実験としてABN
% のAttention mapを，ABNが正しく認識するようにAttention mapを人手で修正する．
% このAttention mapの修正による挙動の変化を調査することで，人の介入によるABNの有効性を検証する．

\begin{figure}[t]
\centering
\includegraphics[width=1.0\linewidth]{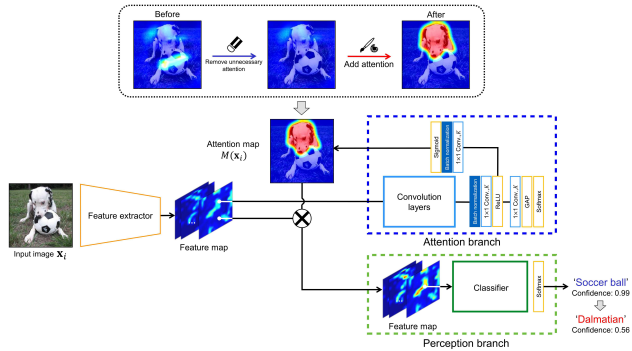}
\caption{Editing procedure of an attention map.}
\label{fig:edit_attention_map}
\end{figure}

\begin{table}[t]
\centering
{\footnotesize \tabcolsep = 3mm
\begin{tabular}{lcc}
% No. of valid. sample & \multicolumn{2}{c}{1$k$} \\ \hline 
& top-1 & top-5 \\ \hline \hline
Before editing & 100.0 & 19.0 \\
After editing & {\bf 83.2} & {\bf 15.8} \\
\end{tabular}
}
\vspace{1mm}
\caption{Top-1 and top-5 errors by edited attention map on validation samples on ImageNet dataset (1k) [$\%$].}
\label{tab:error_attention_edit}
\end{table}

\begin{figure}[t]
\centering
\includegraphics[width=1.0\linewidth]{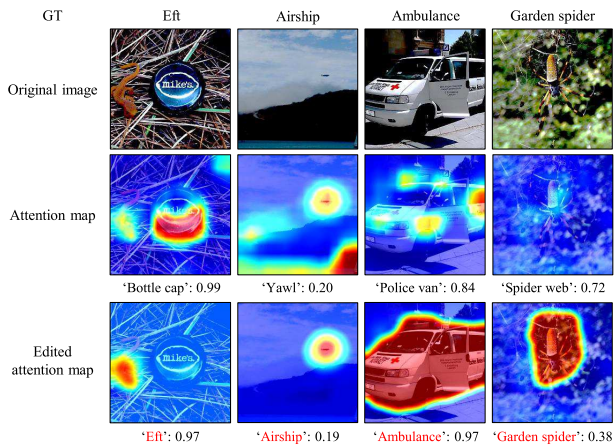}
\caption{Example of conventional and edited attention maps.}
\label{fig:revise_attention_map}
\end{figure}

\subsection{Editing of attention map}
% 検証実験では，ImageNet Dataset~\cite{Deng2009}のvalidationサンプルを用いて，修正したAttention mapをABNに与え，推論後の認識結果の変動を調査する．
% 検証に使用するモデルは，ResNetの152層モデルにABNを導入したモデルである~(以下，ResNet152+ABN)．
% ResNet152+ABNは，ImageNet Datasetの120万枚の学習サンプルを用いて学習し，検証ではResNet152+ABNが誤認識したサンプルから1,000サンプルを選択し，調整に使用する．

In this experiment, we used an ABN whose backbone is 152-layer ResNet \cite{He2016} (ResNet152+ABN) as a network model. ResNet152+ABN is trained with $1,200$k training samples from the ImageNet dataset. Then, we selected the $1$k misclassified samples from the validation samples and edited their attention maps.

% We use validation samples of the ImageNet dataset.
% We replace the attention map during inference with the modified attention map and then check the changes of the classification results.
% As a network model, we use ResNet, which consists of 152 layers, along with the ABN~(ResNet152+ABN).
% ResNet152+ABN is trained with $1,200k$ training samples from the ImageNet dataset.
% Then, we select the $1k$ mis-classified samples from the validation samples and modify these attention maps.

\begin{figure*}[t]
\centering
\includegraphics[width=0.95\linewidth]{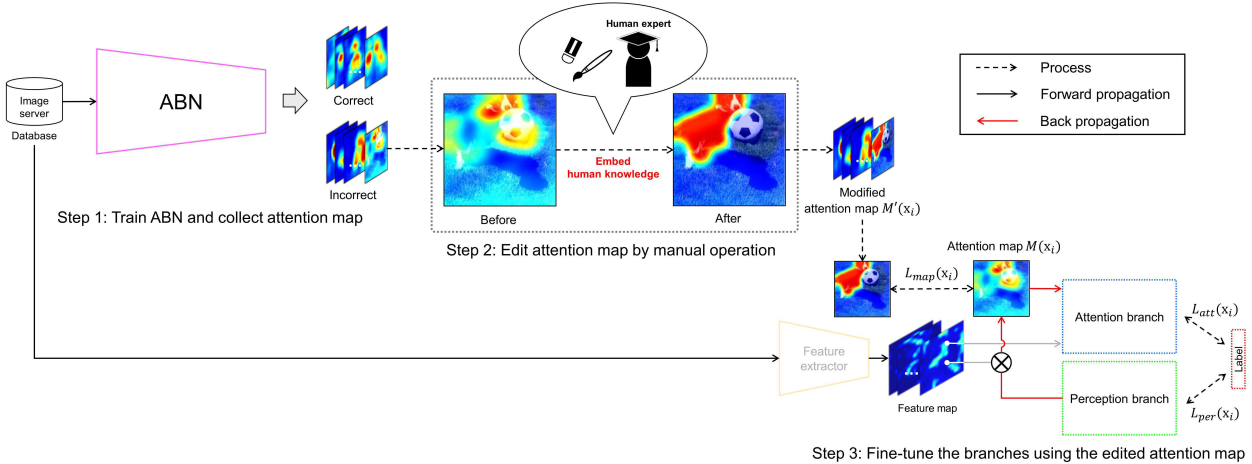}
\caption{Process flow of the proposed method.}
\label{fig:fine_tuning_ABN}
\end{figure*}

% trainサンプル1,281,167枚とvalidationサンプル 50,000枚
% 図~\ref{fig:edit_attention_map}に，Attention mapの修正方法について示す．
% はじめに，誤認識したサンプルをResNet152+ABNに入力し，Perception branchから認識結果を，Attention branchからAttention mapを獲得する．
% ここで，得られるAttention mapは14$\times$14 pixelsである．
% 次に，Attention mapを修正するために，Attention mapを224$\times$224に拡大し，Attention mapを手動により修正する．
% 図~\ref{fig:edit_attention_map}の例の場合，Ground truth~(GT)が``Dalmatian"である画像をResNet152+ABNへ入力した時，2つの物体が画像中に含まれているため``Soccer ball"と誤認識した．
% このとき，Attention mapを可視化すると，``Soccer ball"に注視していることがわかる．
% そこで，注視している領域を``Soccer ball"から``Dalmatian"に手動で変更する．
% この修正により，ResNet152+ABNの認識結果は``Soccer ball"から``Dalmatian"に変化することがわかる．

% The procedure for modifying the attention map is shown in Fig.~\ref{fig:edit_attention_map}.
Figure~\ref{fig:edit_attention_map} shows the editing procedure of an attention map. We first input a misclassified sample to ResNet152+ABN and obtain the attention map from the attention branch, where the size of the attention map is $14 \times 14$ pixels. Then, we edit the obtained attention map manually. Note that the attention map is resized to $224 \times 224$ pixels and is overlaid with the input image for ease of manual editing.
% Second, to edit the obtained attention map, it is resized to $224 \times 224$ pixels and modified manually.
The edited attention map is resized to $14 \times 14$ pixels and used for an attention mechanism to infer classification results from the perception branch.
% todo: 具体的なダルメシアンやサッカーボールの話はなくても良いのでは？（or もっと短く）
In the example shown in Fig.~\ref{fig:edit_attention_map}, the attention map obtained from ResNet152+ABN classifies the input image as ``Soccer ball'' and also highlights the corresponding object. By editing the attention map to highlight ``Dalmatian" and using it for the attention mechanism, the classification result is successfully adjusted to ``Dalmatian".

% As an example, if we input the image shown in Fig.~\ref{fig:edit_attention_map} whose GT is ``Dalmatian" to ResNet152+ABN, it is mis-classified as ``Soccer ball" because two objects are included in the input image.
% Therefore, when we visualize the attention map, ABN pays attention to ``Soccer ball".
% We manually modify the attention region from ``Soccer ball" to ``Dalmatian", as shown in Fig.~\ref{fig:edit_attention_map}, and as a result, the recognition result of ResNet152+ABN is changed from ``Soccer ball" to ``Dalmatian".

% Attention mapの修正例を図~\ref{fig:revise_attention_map}に示す．
% ImageNet Datasetでは，図~\ref{fig:revise_attention_map}の例のように一枚の画像に複数のカテゴリの物体が写っている場合がある．
% そのため，左2列の例のAttention mapのように，それぞれの物体に対して注視することで誤認識を誘発する．
% 1列目の例では，GTが``Eft"であるのに対し，``Eft"の横に存在する``Bottle cap"に対してAttention mapが強く反応し，誤認識している．
% この例に対して，``Bottle cap"のAttention mapを取り除くことで認識結果が``Eft"に変化してることがわかる．
% 2列目の例においても，``Airship"と``Yawl"に対してAttention mapが強く反応した上で誤認識している．
% そのため，``Yawl"の注視領域を取り除くことで，``Airship"と認識結果を調整できることがわかる．

Examples of the edited attention map are provided in Fig.~\ref{fig:revise_attention_map}. In the two left columns, images contain objects from multiple categories and ResNet152+ABN misclassifies these images due to focusing on different objects. 
% Some of the ImageNet samples include objects from multiple categories in an image, and ResNet152+ABN mis-classifies due to focusing on different objects, as shown in the examples in the two left columns in the figure.
For example, in the first column, although the GT is ``Eft", ResNet152+ABN recognizes ``Bottle cap", because the attention map highlights the ``Bottle cap". By removing the attention region of ``Bottle cap'' and adding the attention to ``Eft", the recognition result of ABN is changed to ``Eft''. In the second column, ResNet152+ABN also misclassifies to ``Yawl" because the attention map highlights both ``Airship" and ``Yawl".
By removing the attention location of ``Yawl", we can adjust the recognition result to ``Airship".
Meanwhile, in the two right columns, the attention maps do not highlight the entire objects and incorrect classification results are provided. By editing the attention maps to highlight the entire objects, the classification results are adjusted correctly.

% Attention mapの修正によるtop-1とtop-5 errorの変化を表~\ref{tab:error_attention_edit}に示す．
% ここで，検証に使用する1,000サンプルは上位一位の認識結果が誤認識したサンプルから収集したため，誤認識率は100~$\%$である．
% 表~\ref{tab:error_attention_edit}の結果から，Attention mapの修正により16.8~$\%$の誤認識を削減したことがわかる．
% top-5 errorにおいても，Attention mapの修正により19.0~$\%$から15.8$\%$までtop-5 errorを削減できた．

We show the top-1 and top-5 errors before and after editing the attention map in Tab.~\ref{tab:error_attention_edit}. Here, the top-1 error before editing is $100\%$ because the 1,000 validation samples we used are collected from false recognition on the top-1 recognition result. We can reduce the top-1 error by $16.8\%$ by editing the attention maps. In the top-5 error, we can also reduce from 19.0$\%$ to 15.8$\%$.

\begin{figure}[tb]
\centering
\includegraphics[width=1.0\linewidth]{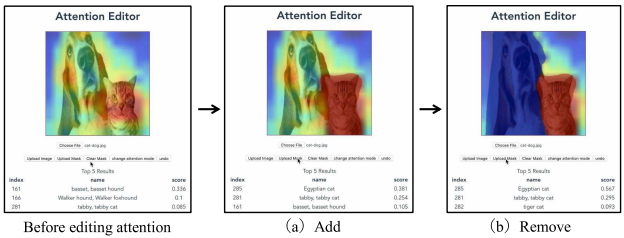}
\caption{Attention map editor tool. (a) Addition of attention. (b) Removal of attention.}
\label{fig:attention_editor}
\end{figure}

%%%%%%%%%%%%%%%%%%%%%%%%%%%%%%%%%%%%%%%%%%%%%%%%%%%%%%%%%%%%%%%%%%%%%%%%
% Proposed method
%%%%%%%%%%%%%%%%%%%%%%%%%%%%%%%%%%%%%%%%%%%%%%%%%%%%%%%%%%%%%%%%%%%%%%%%
\section{Proposed method}
In this section, we discuss how to embedding human knowledge into deep neural networks.
The results discussed in Sec.~\ref{sec:edit_attention} demonstrate that the recognition result of ABN can be adjusted by editing the attention map. This suggests that ABN can be applied to embedding human knowledge into the network. Therefore, we propose fine-tuning the attention and perception branches of ABN by using the edited attention map. By training the attention and perception branches with the edited attention map including human knowledge, ABN can output an attention map that considers this knowledge and thereby improve the classification performance.
%\ref{sec:edit_attention}章の検証実験から，ABNのAttention mapは修正に合わせて認識結果を変更できる性質を持つことが判明した．
%この検証実験から，ABNは深層学習におけるHITLに応用可能であることがわかる．
%そこで本研究では，人の知見により修正したAttention mapを用いてABNをファインチューニングする．
%ファインチューニングにより人の知見を取り入れたAttention mapを学習することで，ABNが人の知見を考慮したAttention mapを出力できるようになる．

\begin{figure*}[tb]
\centering
\includegraphics[width=1.0\linewidth]{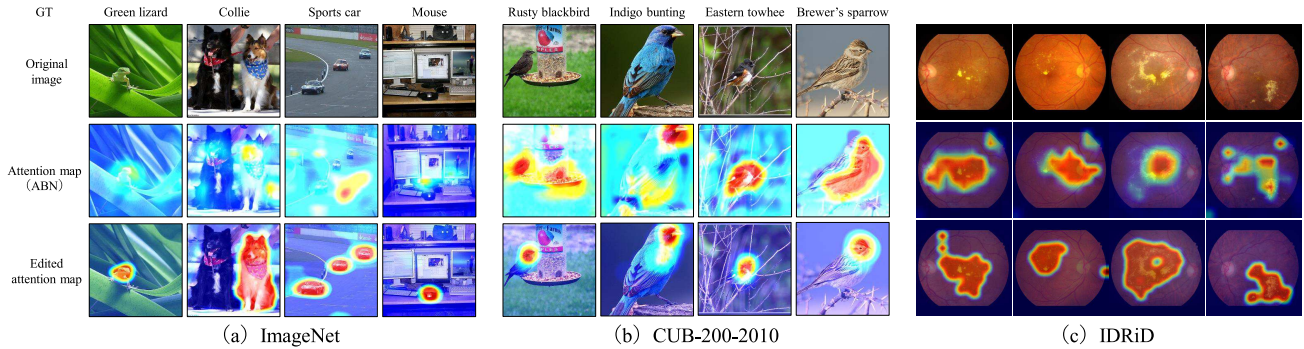}
\caption{Example of edited attention map for each dataset.}
\label{fig:attention_map_revise}
\end{figure*}

% \subsection{ABN with HITL}
\subsection{Embedding human knowledge via edited attention map}
The process flow of the proposed method is shown in Fig.~\ref{fig:fine_tuning_ABN}. First, an ABN model is trained using training samples with labels, and then we collect the attention maps of these samples from the trained model. Here, we only collect the attention maps of misclassified training samples. Second, we edit each of the attention maps based on human knowledge to recognize them correctly. %The edited attention maps include human knowledge.
Third, the attention and perception branches of ABN are fine-tuned with the edited attention maps. During the fine-tuning process, we update the parameters of the attention and perception branches by using the loss calculated from the attention map obtained from ABN and the edited attention map in addition to the loss of ABN (the details are described in Sec.~\ref{sec:loss}).

%提案手法である深層学習ベースのHITLの処理の流れを図~\ref{fig:fine_tuning_ABN}に示す．
%はじめに，学習データと教師ラベルを基にABNを学習し，学習済みのABNを用いてAttention mapを収集する．
%ここで，Attention mapの収集では，誤認識した学習サンプルのAttention mapのみを収集する．
%次に，収集したAttention mapは，正確に認識できるようにAttetion mapを人の知見によって修正する．
%その後，修正したAttention mapを用いてABNのAttention branchとPerception branchをファインチューニングする．
%ファインチューニングする際，ABNのAttention branchとPerception branchの学習誤差に加えて，出力されたAttention mapと修正したAttention mapを用いて学習誤差を算出し，ファインチューニングする．

\subsection{Manual edit of attention map}

Depending on the dataset, existing of additional information, and expertise for specific applications, several editing approaches can be considered. Herein, we introduce the three methods we use to edit attention maps.

{\bf ImageNet dataset}\hspace{3mm}
We manually edit the attention maps of the ImageNet dataset with the same process as described in Sec.~\ref{sec:edit_attention}. To edit as many attention maps as possible, we created a tool that can edit attention maps interactively, as shown in Fig.~\ref{fig:attention_editor}. This tool can add (Fig.~\ref{fig:attention_editor}(a)) and remove (Fig.~\ref{fig:attention_editor}(b)) an attention region simply by dragging the mouse. With this tool, we can edit attention maps interactively while verifying the top-3 classification results. Examples of the edited attention maps are shown in Fig.~\ref{fig:attention_map_revise}(a). These maps are edited so that an object or characteristic region with respect to the GT is highlighted \footnote{The edited attention maps will be released. See \url{https://attention-editor.netlify.com/} for the web page of this tool.}.

%ImageNet DatasetにおけるAttention mapの修正では，\ref{sec:edit_attention}章の検証実験と同様の処理で修正を行う．
%より多くのAttention mapを修正するために，図~\ref{fig:attention_editor}のようなAttention mapを修正するツール\footnote{本ツールのWebページと修正したAttention mapのデータは,公開予定である.}を作成した．
%このツールは，マウスカーソルのドラッグ$\&$ドロップに合わせて，Attention mapの注視領域の追加(図~\ref{fig:attention_editor}(a))と除去(図~\ref{fig:attention_editor}(b))する．
%このとき，推論結果~(top-3)を確認しながらAttention mapをインタラクティブに修正する．
%図~\ref{fig:attention_editor}のツールにより修正したAttention mapを，図~\ref{fig:attention_map_revise}(a)に示す．
%図~\ref{fig:attention_map_revise}(a)のように，特定のカテゴリに対する物体領域や，ある物体の特有の特徴的な領域に対して強く反応するようにAttention mapが修正されている．

{\bf CUB-200-2010 dataset}\hspace{3mm}
In the CUB-200-2010 dataset~\cite{WelinderEtal2010}, we embed human knowledge into an attention map by using bubble information \cite{deng2013fine}. The bubble information represents the attention region by means of the position and scale of the circular bounding box when multiple users distinguish two categories of birds. This information is an important human knowledge to recognize the multiple categories of birds. For this reason, we make an attention map with human knowledge from the bubble information.

%CUB-200-2010 Datasetでは，人の知見により付与されたbubble情報を用いてAttention mapを作成する．
%bubble情報は，複数のユーザによって付与された情報であり，2種類の鳥を認識する際に，注視した領域を複数の円の位置とスケールで表現した情報である．
%この情報は，複数種の鳥を高精度に認識するために必要な人の知見となる．
%そのため，このbubble情報を用いてAttention mapを作成する．

\begin{figure}[tb]
\centering
\includegraphics[width=0.95\linewidth]{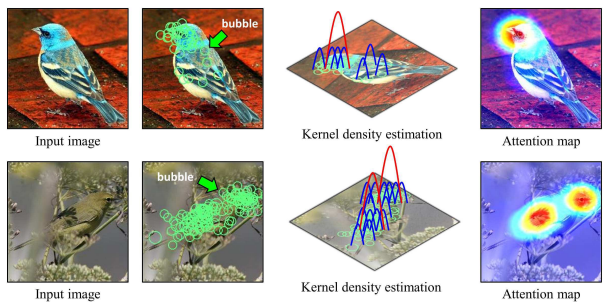}
\caption{Making an attention map from the bubble by kernel density estimation.}
\label{fig:bubble}
\end{figure}

For each bird image, bubbles are annotated by multiple users. The number of bubbles given by one user is not limited. To make an attention map from the bubbles, we use a kernel density estimation with multiple bubbles, as shown in Fig.~\ref{fig:bubble}. A dense region of bubbles indicates an important region for recognizing the bird category. The density of bubble information enables us to obtain the attention map embedded with human knowledge, as shown in Fig.~\ref{fig:attention_map_revise}(b). The map is then normalized to [0-1] and used for the proposed fine-tuning method.

%bubbleは，1つのサンプルに対して複数ユーザのbubbleが付与されており，1ユーザが付与するbubbleの数に制限は設けられていない．
%bubbleからAttention mapを作成するために，図~\ref{fig:bubble}のようにサンプルに付与されている複数のbubbleからカーネル密度推定を行う．
%bubbleの密度が高い領域は鳥類を認識する上で重要な領域を示しているため，図~\ref{fig:bubble}に示すAttention mapを得ることができる．
%その後，Attention mapの値を[0-1]に正規化することで，ファインチューニングに応用する．

{\bf Fundus image dataset (IDRiD)}\hspace{3mm}
To achieve an automatic diagnosis, medical image recognition has been attempted for various recognition tasks, such as retinal disease recognition~\cite{Fauw2018} and risk forecasting of heart disease~\cite{Poplin2018}. In actual medical practice, a system that can explain the reason behind a decision is required in order to enhance the reliability of the diagnosis. The presentation of decision-making in automatic diagnosis is attracting considerable attention because automatic diagnosis is greatly helpful to doctors when making a diagnosis. In this paper, we evaluate the disease recognition of a fundus image.

%医用画像認識では，眼底画像から網膜疾患の識別\cite{Fauw2018}や，心疾患リスクの予測\cite{Poplin2018}など，自動診断の実現に向けた様々な試みが行われている．
%実際の医療現場においては，自動診断における患者の信頼性を高めるために，診断根拠を説明するシステムが必要となる．
%自動診断における根拠の提示は，医師による診断をサポートできるため，注目されている技術である．
%本稿では，自動診断における医用画像認識を対象とし，眼底画像を用いた疾患識別を行う．

For this disease recognition, we use the Indian Diabetic Retinopathy Image Dataset (IDRiD) \cite{Prasanna2018}. IDRiD is concerned with the disease grade recognition of retina images, and the presence or absence of diseases is recognized from exudates and hemorrhages. IDRiD includes a segmentation label of disease regions annotated by a specialist, as shown in Fig.~\ref{fig:attention_map_revise}(c). We edit the attention map of the disease classification task by using the segmentation label.

%本研究では，眼底画像を用いた疾患識別を対象とし，Indian Diabetic Retinopathy Image Dataset~(IDRiD)~\cite{Prasanna2018}を用いる．
%IDRiDは眼底画像による疾患のグレード識別を対象としており，白斑や出血の有無で疾患の識別を行う．
%IDRiDには，図~\ref{fig:attention_map_revise}(c)のように専門家の医師の知見でラベル付けされた，疾患領域のセグメンテーションラベルがある．
%疾患識別におけるAttention mapには，セグメンテーションラベルを用いて修正する．

\subsection{Fine-tuning of the branches}\label{sec:loss}

After editing attention maps including human knowledge, ABN is fine-tuned with these maps. 
% In the proposed fine-tuning method, we add a loss $L_{map}$ to the conventional ABN loss function, which is defined as
In the proposed fine-tuning method, we formulate the loss function $L$ in addition to the conventional ABN loss function $L_{abn}$. 
Let ${\bf x}_i$ be the $i$-th sample in the training dataset. 
The loss function of ABN is calculated by
\begin{equation}
L_{abn}({\bf x}_i)=L_{att}({\bf x}_i)+L_{per}({\bf x}_i),
\label{eq:conv_abn_loss}
\end{equation}
where $L_{att}$ and $L_{per}$ are conventional cross entropy losses for the attention and perception branches, respectively.
The loss function of the fine-tuning $L({\bf x}_i)$ is defined as
\begin{equation}
L({\bf x}_i) = L_{abn}({\bf x}_i) + {L}_{map}({\bf x}_i).
\label{eq:proposed_training_loss}
\end{equation}

As the loss of the attention maps $L_{map}$, we use the L2 norm between the two attention maps. Here, we denote an output attention map from ABN and a edited attention map as $M({\bf x}_i)$ and $M'({\bf x}_i)$, respectively. The attention map loss~$L_{map}$ are formulated by
\begin{equation}
L_{map}({\bf x}_i)=\gamma \| M'({\bf x}_i) - M({\bf x}_i) \|_{2},
\label{eq:lmap_loss}
\end{equation}
where $\gamma$ is a scale factor. Typically, $L_{map}$ is larger than $L_{att}$ and $L_{per}$. Hence, we adjust the effect of $L_{map}({\bf x}_i)$ by scaling $L_{map}$ with $\gamma$.

By introducing $L_{map}$, ABN is optimized so that an output attention map is close to the edited attention map including human knowledge. In this way, we can embed human knowledge into a network via the edited attention map. 
% In this way, ABN is fine-tuned to output an attention map corresponding to human knowledge.
During the fine-tuning, the proposed method optimizes the attention and perception branches of ABN. The feature extractor that extracts the feature map from an input image is not updated during the fine-tuning process.

\section{Experiments}

\subsection{Experimental details}

We evaluate the proposed method on image classification~\cite{Deng2009}, fine-grained recognition~\cite{WelinderEtal2010}, and fundus image classification~\cite{Prasanna2018} tasks. 
% by using the ImageNet dataset~\cite{Deng2009}, CUB-200-2010 dataset~\cite{WelinderEtal2010}, and IDRiD~\cite{Prasanna2018}, respectively.
Also, in order to quantitatively evaluate the explanation capability of the attention map, we use the deletion metric, the insertion metric, and the degree of similarity between the edited attention map and the attention map output by the network.
% todo: 評価指標の参考文献追加（あと文章を整理）

%評価実験では，ImageNet Dataset~\cite{Deng2009}，CUB-200-2010 Dataset~\cite{WelinderEtal2010}，IDRiD~\cite{Prasanna2018}を用いて，一般物体認識，詳細認識，医用画像認識における提案手法の有効性の検証を行う．
%また，Attention mapの説明性を数値的に評価するためにDeletion metric，Insertion metric，修正したAttention mapとネットワークが出力したAttention mapとの類似度を用いる．

\begin{table}[t]
\centering
{\footnotesize \tabcolsep = 2.4mm
\begin{tabular}{llr}
No. of categories & Model & top-1 error \\ \hline \hline 
\multirow{9}{*}{\shortstack{10 random \\categories}} & ResNet18 & 9.00\hspace{3.5mm} \\
& ResNet34 & 9.60\hspace{3.5mm} \\
& ResNet50 & 12.00\hspace{3.5mm} \\ \cline{2-3}
& ResNet18+ABN & 8.40\hspace{3.5mm} \\
& ResNet34+ABN & 7.60\hspace{3.5mm} \\
& ResNet50+ABN & 11.20\hspace{3.5mm} \\ \cline{2-3}
& Proposed (ResNet18+ABN) & {\bf6.20}\hspace{3.5mm} \\
& Proposed (ResNet34+ABN) & {\bf7.40}\hspace{3.5mm} \\
& Proposed (ResNet50+ABN) & {\bf10.80}\hspace{3.5mm} \\ \hline
\multirow{3}{*}{\shortstack{100 worst \\categories}} & ResNet152 & 31.90\hspace{3.5mm} \\
& ResNet152+SE+ABN & 31.16\hspace{3.5mm} \\
& Proposed (ResNet152+SE+ABN) & {\bf30.88}\hspace{3.5mm} \\
\end{tabular}
}
\vspace{1mm}
\caption{Top-1 error rates of the conventional and proposed methods on ImageNet dataset [$\%$].}
\label{tab:comp_benchmark_imagenet}
\end{table}
\begin{figure}[t]
\centering
\includegraphics[width=0.95\linewidth]{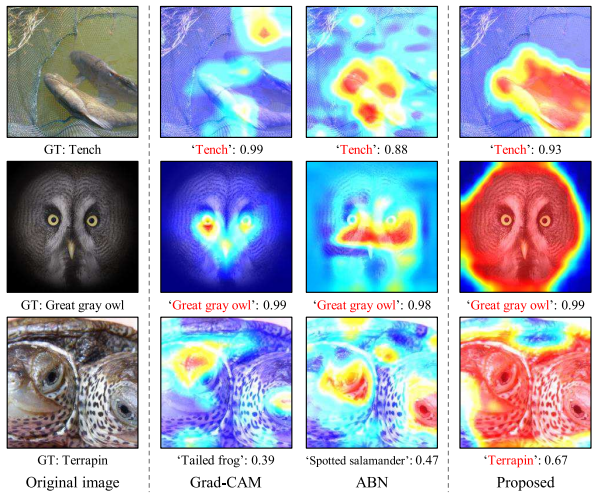}
\caption{Examples of conventional and proposed attention maps on ImageNet dataset.}
\label{fig:imagenet_attentoin_map}
\end{figure}

\begin{table}[!h]
\centering
{\footnotesize \tabcolsep = 2.4mm
\begin{tabular}{lcccc}
& \multicolumn{2}{c}{scratch} & \multicolumn{2}{c}{pre-trained}\\
\cline{2-5}
Model & top-1 & top-5 & top-1 & top-5 \\ \hline \hline
Deng’s~method~\cite{deng2013fine} & 32.80 & --  & -- & -- \\ \hline
ResNet18 & 28.38 & 52.62 & 62.58 & 83.25 \\
ResNet34 & 27.39 & 53.28 & 67.59 & 85.13 \\
ResNet50 & 28.02 & 54.33 & 69.27 & 88.39 \\ \hline
ResNet18+ABN & 32.38 & 57.27 & 63.57 & 83.45 \\
ResNet34+ABN & 30.99 & 53.68 & 68.25 & 87.73 \\
ResNet50+ABN & 31.68 & 57.01 & 71.68 & 89.09 \\ \hline
Proposed (ResNet18+ABN) & {\bf36.96} &  {\bf61.66} & {\bf64.72} & {\bf83.71} \\
Proposed (ResNet34+ABN) & {\bf38.15} & {\bf62.78} & {\bf69.27} & {\bf87.88} \\
Proposed (ResNet50+ABN) & {\bf37.42} & {\bf62.08} & {\bf72.07} & {\bf90.37} \\
\end{tabular}
}
\vspace{1mm}
\caption{Top-1 and top-5 errors of conventional and proposed methods on CUB-200-2010 dataset~[$\%$].}
\label{tab:comp_benchmark_cub}
\end{table}

\begin{figure}[!h]
\centering
\includegraphics[width=0.95\linewidth]{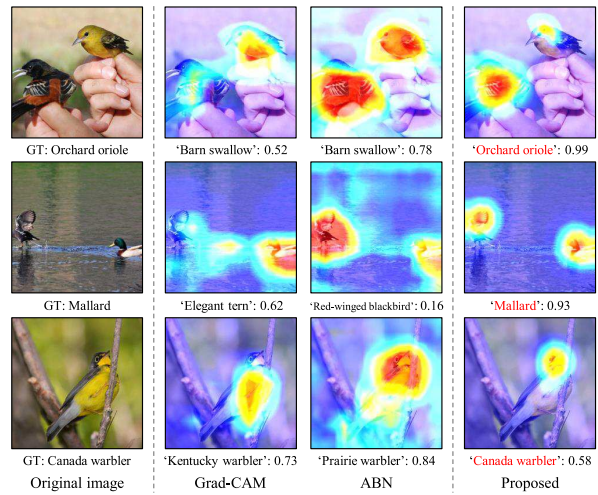}
\caption{Examples of conventional and proposed attention maps on CUB-200-2010 dataset.}
\label{fig:bubble_att_per}
\end{figure}

{\bf ImageNet dataset}\hspace{3mm}
We collect misclassified training samples of the top-1 result in the ImageNet dataset to edit the attention map and to use for fine-tuning. 
% The target samples are the mis-classified samples of the top-1 result in the training samples in ImageNet dataset.
We edit the attention maps of 100 categories with lower classification performance to evaluate the improvement and randomly selected ten categories among them. The number of edited attention maps was 30,917, and editing was performed by 43 users. During fine-tuning, we compare two training performances: training only ten categories and training all 100 categories of the edited attention maps.
 
%ImageNet DatasetにおけるAttention mapの修正する対象は，ImageNet Datasetの学習サンプルであり，上位一位における認識結果が誤認識したサンプルを対象とする．
%加えて，ランダムにソーティングした10カテゴリと誤認識したサンプル数の割合が多い上位100位のカテゴリを対象に評価する．
%修正したAttention mapの枚数は30,917枚であり，約40人のユーザで修正した．
%ファインチューニング時は，修正した10カテゴリのみ使用する場合と修正した100カテゴリのみ用いる場合の2種類を行い，比較する．

Our baseline models are ResNet18, ResNet34, ResNet50, and ResNet152 that includes a SENet~\cite{Hu2017}. We used the same learning conditions as~\cite{fukui2018attention}.

%本実験では，ResNet18，ResNet34，ResNet50にABNを導入したモデル~(以下，ResNet18+ABN，ResNet34+ABN，ResNet50+ABN)とResNet152にSqueeze-and-Excitation~Network~\cite{Hu2017}とABNを導入したモデル~(以下，ResNet152+SE+ABN)を対象とする.
%これらのモデルの学習条件は文献\cite{fukui2018attention}と同様である．

%これらのモデルは，Stocastic Gradient Decent~(SGD) with Momentumにより90~epoch学習しており，
%Data augmentationの方法は文献\cite{fukui2018attention}と同様である．
%ファインチューニング時はStocastic Gradient Decent~(SGD) with Momentumにより90~epoch学習し，31~epochと61~epochでlearning rate dropにより学習率を0.1倍する．
%バッチサイズは10カテゴリのみの場合は16，100カテゴリのみの場合は128であり，Data augmentationの方法は文献\cite{fukui2018attention}と同様である．

\begin{table*}[tb]
\centering
{\footnotesize \tabcolsep = 3mm
\begin{tabular}{lcccccc}
& \multicolumn{2}{c}{ResNet18} & \multicolumn{2}{c}{ResNet34} & \multicolumn{2}{c}{ResNet50} \\
\cline{2-7}
Method & Deletion & Insertion & Deletion & Insertion & Deletion & Insertion \\ \hline \hline
Grad-CAM~\cite{Selvaraju2017} & {\bf0.0719} & 0.3729 & 0.1329 & 0.4784 & 0.1480 & 0.3973 \\
ABN~\cite{fukui2018attention} & 0.1560 & 0.3633 & 0.0920 & 0.5734 & 0.0706 & 0.6330 \\
Proposed & 0.1039 & {\bf0.4995} & {\bf0.0778} & {\bf0.5842} & {\bf0.0564} & {\bf0.6828} \\
\end{tabular}
}
\vspace{1mm}
%\caption{CUB200-2010 Datasetにおける従来の視覚的説明と提案手法のAttention mapのDeletion (lower is better)とInsertion (higher is better)スコアによる比較}
\caption{Comparison of the deletion (lower is better) and insertion (higher is better) scores of conventional visual explanation and proposed method on CUB-200-2010 dataset.}
\label{tab:cub_del_in}
\end{table*}

{\bf CUB-200-2010 dataset}\hspace{3mm}
The CUB-200-2010 dataset includes attention maps created by the bubble information for all training samples. Therefore, the training samples are sorted by their confidence, and the samples are used for fine-tuning from the lowest confidence.

% Then, gradually increase the number of training samples to be modified according to the training results as in active learning \cite{angluin1988queries,atlas1990training,lewis1994sequential}, and repeat Step 2 and Step 3 as shown in Fig.~\ref{fig:fine_tuning_ABN}.

%CUB-200-2010 Datasetはサンプル数が少ないため，過学習を引き起こしやすく，scratchから学習させることが困難である．
%そのため，scratchから学習する場合とImageNetの事前学習モデルを使用する場合の2つの学習方法で検証する．
%CUB-200-2010 Datasetでは，全学習サンプルに対してbubble情報により作成したAttention mapがある．
%そのため，学習サンプルをconfidenceの低い順にソーティングし，confidenceの低いサンプルからファインチューニングに用いる．
%その後，Active Learning \cite{angluin1988queries,atlas1990training,lewis1994sequential}のように学習結果に応じて修正する学習サンプル数を徐々に増やし，図~\ref{fig:fine_tuning_ABN}のようにStep2とStep3を繰り返す．

Since the CUB-200-2010 dataset has a small number of samples, it is easy for over-fitting to occur, and learning from scratch is difficult. For this reason, we evaluated two learning methods: training from scratch and fine-tuning the pre-trained model on ImageNet. These models are trained by SGD with momentum in 300 epochs. The learning rate is decreased to 0.1 times at 150 and 225 epochs. The mini-batch size is 16.

%これらのモデルは，SGD with Momentumにより300~epoch学習する．
%ここで，150~epochと225~epochでlearning rate dropにより0.1倍する．
%学習パラメータはバッチサイズを16，学習の更新回数を300~epochとする．
%Data augmentationはImageNet Datasetと同様の処理で行う．

\begin{table}[tb]
\centering
{\footnotesize \tabcolsep = 2.4mm
\begin{tabular}{lccc}
% & \multicolumn{3}{c}{Mean square error} \\ \cline{2-4}
Method & ResNet18 & ResNet34 & ResNet50 \\ \hline \hline
CAM~\cite{Zhou2016} & 0.6456 & 0.7658 & 0.6031 \\
Grad-CAM~\cite{Selvaraju2017} & 0.4502 & 0.4831 & 0.3875 \\
ABN~\cite{fukui2018attention} & 0.1682 & 0.3022 & 0.5499 \\
Proposed & {\bf0.1136} & {\bf0.1597} & {\bf0.2049} \\
\end{tabular}
}
\vspace{1mm}
\caption{Comparison of the similarity of the attention map by mean square error of the conventional visual explanation and proposed method on CUB-200-2010 dataset.}
%\caption{CUB200-2010 Datasetにおける従来の視覚的説明と提案手法のAttention mapの類似度による比較}
\label{tab:cub_coincidence}
\end{table}

\begin{table}[t]
\centering
{\footnotesize \tabcolsep = 7.0mm
\begin{tabular}{lc}
Model & Accuracy \\ \hline \hline
AlexNet & 89.66  \\ 
ResNet18 & 89.78 \\ 
ResNet34 & 94.44 \\
ResNet50 & 95.83 \\ \hline
AlexNet+ABN & 93.11 \\ 
ResNet18+ABN & 95.33 \\
ResNet34+ABN & 96.88 \\
ResNet50+ABN & 97.22 \\ \hline
Proposed (AlexNet+ABN) & {\bf 96.78} \\
Proposed (ResNet18+ABN) & {\bf96.88} \\
Proposed (ResNet34+ABN) & {\bf97.23} \\
Proposed (ResNet50+ABN) & {\bf99.17} \\
\end{tabular}
}
\vspace{1mm}
\caption{Comparison of the accuracy of the conventional and the proposed methods on IDRiD~[$\%$].}
\label{tab:comp_retina_recog}
\end{table}

\begin{table*}[t]
\centering
{\footnotesize \tabcolsep = 3mm
\begin{tabular}{lcccccc}
\label{tab:idrid_del_in}
& \multicolumn{2}{c}{ResNet18} & \multicolumn{2}{c}{ResNet34} & \multicolumn{2}{c}{ResNet50} \\ \cline{2-7}
Method & Deletion & Insertion & Deletion & Insertion & Deletion & Insertion \\ \hline \hline
Grad-CAM~\cite{Selvaraju2017} & 0.3058 & 0.6201 & {\bf0.2958} & 0.7085 & {\bf0.2854} & 0.7465 \\
ABN~\cite{fukui2018attention} & 0.4201 & 0.6028 & 0.3900 & 0.6878 & 0.3844 & 0.6524 \\
Proposed & {\bf0.2732} & {\bf0.8704} & 0.2978 & {\bf0.8832} & 0.3102 & {\bf0.8923} \\
\end{tabular}
}
\vspace{1mm}
\caption{Comparison of the deletion (lower is better) and insertion (higher is better) scores of the conventional visual explanation and the proposed method on IDRiD.}
\label{tab:idrid_del_in}
\end{table*}

\begin{table}[tb]
\centering
{\footnotesize \tabcolsep = 2.4mm
\begin{tabular}{lccc}
% & \multicolumn{3}{c}{Mean square error} \\ \cline{2-4}
Method & ResNet18 & ResNet34 & ResNet50 \\ \hline \hline
CAM~\cite{Zhou2016} & 0.3749 & 0.3300 & 0.2287 \\
Grad-CAM~\cite{Selvaraju2017} & 0.1521 & 0.1329 & 0.1532 \\
ABN~\cite{fukui2018attention} & 0.1241 & 0.1309 & 0.1286 \\
Proposed & {\bf0.0893} & {\bf0.0927} & {\bf0.0904} \\
\end{tabular}
}
\vspace{1mm}
\caption{Comparison of the similarity of the attention map by mean square error of the conventional visual explanation and proposed method on IDRiD.}
\label{tab:idrid_coincidence}
\end{table}

{\bf IDRiD}\hspace{3mm}
IDRiD contains 81 diseased images and 120 healthy images based on the existence of hemorrhages, hard exudates, and soft exudates. We create the edited attention maps by using semantic segmentation labels annotated by medical doctors. We evaluate IDRiD by 5-fold cross validation. Our baseline models are an AlexNet, ResNet18, ResNet34 and ResNet50-based CNNs. The networks are trained by SGD with momentum, and the number of training iterations is 9,500 epochs. The batch size is 20 and the size of each image is 360 $\times$ 360 pixels. Data augmentation is as follows: mirroring, intensity change, scaling, and rotation.

% 医用画像診断における評価で用いるIDRiDには，81枚の画像が含まれている．
% 81枚のうち，白斑や出血の有無をもとに疾患画像38枚，正常画像43枚に分割し，本実験ではK-hold cross validationにより評価する．
% 評価に使用するネットワークはAlexNetをベースとしており，学習サンプル数の不足に対処するために畳み込み層のチャンネル数を半分に削減している．
% IDRiDもCUB-200-2010 datasetと同様に全学習サンプルに対して専門家の医師の知見でラベル付けされた，疾患領域のセグメンテーションラベルを用いて修正したAttention mapがある．そのため，学習サンプルをconfidenceの低い順にソーティングし，CUB-200-2010 datasetと同様に学習結果に応じて修正する学習サンプル数を徐々に増やし，Step2とStep3を繰り返す．

% unused
%本実験では，81サンプルの2クラス分類問題として扱うため，サンプル数が少なく，ネットワークの構造の変化による効果は薄い．
%そのため，AlexNetのみでも提案手法の効果は十分に確認できる．
%学習時は，SGD with Momentumにより9500~epoch学習している．
%バッチサイズは20であり，入力画像サイズは360$\times$360である．
%また，学習時はミラーリング，輝度変化，スケーリング，回転のData augmentationを施す．

%{\bf Attention mapの数値的な評価}\hspace{3mm}
{\bf Quantitative evaluation of attention map}\hspace{3mm}
In order to quantitatively evaluate the explainability of the attention map, we use the deletion metric, the insertion metric, and the degree of similarity between the edited attention map and the attention map output by the network. The deletion and insertion metrics are evaluation methods proposed by Petsiuk {\it et al.} \cite{Petsiuk2018rise}, which are based on the concept of literature \cite{fong2017interpretable}. The deletion metric measures the decrease of score by gradually deleting the high attention area of an attention map from the input image. Therefore, a lower score means a higher explanation. On the other hand, the insertion metric measures the increase of score by gradually inserting the high attention area of an attention map in the input image. Therefore, a higher score means a higher explanation. In the evaluation, the degree of similarity between the edited attention map and the attention map output by the network is measured by the mean square error. 
%
% Therefore, the closer error is to 0, the higher similarity to the modified attention map.
% todo: この2文の説明はなくても良いかも（上で暗に示してるし．．．）
% In CUB-200-2010 dataset, it measures the similarity between the attention map created by the bubble information and the attention map output by the network.
% In IDRiD, it measures the degree of similarity between the attention map which was modified using a segmentation label of disease regions annotated by a specialist and the attention map output by the network.
%
A higher similarity (i.e., lower error) means that the attention map focuses on the same area as the human operator and thus successfully embedded human knowledge.

%Attention mapの説明性を数値的に評価するためにDeletion metric，Insertion metric，修正したAttention mapとネットワークが出力したAttention mapとの類似度を用いて評価する．
%Deletion metricとInsertion metricは，Petsiukら\cite {Petsiuk2018rise}が提案する評価方法であり，文献~\cite{fong2017interpretable}の考え方をベースにしている．
%Deletion metricは，Attention mapの注目度の高い領域を入力画像から徐々に削除することによるスコアの低下を測定する．
%したがって，スコアが低いほど説明性が高いことを意味している．
%一方で，Insertion metricは重要度の高い領域を徐々に追加することによるスコアの増加を測定する．
%したがって，スコアが高いほど説明性が高いことを意味している．
%修正したAttention mapとネットワークが出力したAttention mapとの類似度による評価では,平均2乗誤差により類似度を測定する．
%そのため，誤差が0に近いほど修正したAttention mapとの類似性が高いことを示す．
%CUB-200-2010 Datasetでは，bubble情報により作成したAttention mapとネットワークが出力したAttention mapとの類似度を測定する．
%IDRiDでは，専門家の医師の知見でラベル付けされた，疾患領域のセグメンテーションラベル用いて修正したAttention mapとネットワークが出力したAttention mapとの類似度を測定する．
%修正したAttention mapは認識する際に人が注視した領域を示しているため，類似度が高いほど人の注視した領域に近い，説明性の高いAttention mapであることを示している．

%-------------------------------------------------------------------------
\subsection{Image classification on ImageNet}
We evaluated the classification performance by using the ImageNet dataset. As in the previous evaluation, we use the 100 worst categories and randomly select ten categories from among them. The accuracies of the conventional ResNet and the proposed method for ten and 100 categories are listed in Tab.~\ref{tab:comp_benchmark_imagenet}. As shown, the accuracy of the proposed method is higher than that of the conventional ABN.

%はじめに，ImageNet Datasetを用いて評価を行う．
%ここでは，ランダムにソーティングした上位10位のカテゴリと各カテゴリにおける誤認識率を算出してソーティングした，ワースト100位までのカテゴリのみを使用する.
%表~\ref{tab:comp_benchmark_imagenet}に，10カテゴリと100カテゴリそれぞれの従来法であるResNetモデルのABNとのエラー率を比較する．
%表~\ref{tab:comp_benchmark_imagenet}から，提案手法の導入によりABNのモデルの精度が向上していることがわかる.

The attention maps of the conventional and proposed methods are shown in Fig.~\ref{fig:imagenet_attentoin_map}. The attention map of the conventional ABN is noisy or focuses on different objects, which results in wrong classifications. In contrast, the proposed method can obtain a clear attention map that highlights the target category object, thus improving the classification performance.

% 従来のABNのAttention mapと提案手法のAttention mapを図~\ref{fig:imagenet_attentoin_map}に示す．
% 従来のABNが出力したAttention mapは，異なる領域に注視したり，ノイズのような注視領域が発生している．
% このようなAttention mapが原因で誤認識が発生している．
% 提案手法のAttentioin mapの可視化では，対象の物体に注視領域が得られていることがわかる．
% また，認識対象の物体ではない人物に対する注視領域も発生していない．
% 加えて，Attention mapの改善により認識結果の改善も確認できた．

\subsection{Fine-grained recognition on CUB-200-2010}
We compared the accuracies of Deng {\it et al}, the conventional ResNet, and the proposed method for top-1 and top-5 accuracy. The results are shown in Tab.~\ref{tab:comp_benchmark_cub}. 
The performances of the conventional ABN trained from scratch and the Deng methods are the same. By fine-tuning the ABN using an attention map with human knowledge, the top-1 accuracies are improved from  4$\%$ to 7$\%$ in the case of scratch. Also, in the case of the pre-trained model on ImageNet, the accuracies are improved by about 1$\%$.
% about top-5 accuracy
Similarly, in the recognition accuracy of top-5, the recognition rate improved from about 4$\%$ to 9$\%$ in the case of scratch, and about 1$\%$ for the pre-trained model on ImageNet.

%表~\ref{tab:comp_benchmark_cub}に，Deng. method，従来のABNと提案手法のtop-1~accuracyとtop-5~accuracyの比較を示す．
%従来のABNでは，Deng. methodと同等の精度であった．
%提案手法のファインチューニングを導入することで，scratchの場合では，top-1の認識精度が約7~$\%$，ImageNetの学習済みモデルを用いた場合では，約1~$\%$向上した．
%top-5の認識精度においても，scratchの場合では認識率が約4~$\%$から9~$\%$，ImageNetの学習済みモデルを用いた場合では約1~$\%$向上した．

Examples of the obtained attention map on the fine-grained recognition are shown in Fig.~\ref{fig:bubble_att_per}. The conventional ABN highlights the entire body of the bird. In contrast, the proposed method highlights the local characteristic regions, such as the color and the head of the bird. In addition, the proposed method removes noise from the attention map by fine-tuning. Thus, the proposed method can also improve the performance of fine-grained recognition.

% CUB-200-2010 datasetにおけるAttention mapの可視化例を，図~\ref{fig:bubble_att_per}に示す．
% 従来のABNでは，鳥の全身に対してAttention mapが強く反応している．
% 一方，提案手法は鳥の判別に有効な特徴的な局所領域のみにAttention mapが反応している．
% また，従来のABNで発生していたノイズのような注視領域が，提案手法により除去されていることがわかる．
% このように，詳細識別において有効なAttention mapを獲得できたため，認識精度を改善できたと考えられる．

For the quantitative evaluation on the explainability of the attention map, we show the deletion (lower is better) and insertion (higher is better) scores of the conventional and proposed methods for test samples of CUB-200-2010 dataset in Tab.~\ref{tab:cub_del_in}. 
% todo: 全体として考察をまとめた
As shown, the proposed method has higher scores than the other methods. In other words, the proposed method provides the clearest visual explanation of all the methods.
% It can be confirmed that the proposed method obtained better evaluation value than ABN from Table~\ref{tab:cub_del_in}.
% Therefore, it can be seen that attention maps with higher explanation than the conventional ABN is obtained.
% In addition, comparing Grad-CAM with the proposed method, it can be confirmed that ResNet34 and ResNet50 have obtained good evaluation values.
%
% todo: grad-CAMとの比較をするなら，attention mapの例も欲しい（ひとまず全部コメントアウト）
% However, the deletion score in ResNet18 was better for Grad-CAM than the proposed method.
% Grad-CAM has a strong tendency to gaze at the entire object to be recognized.
% On the other hand, the ABN gaze area is more local than Grad-CAM.
% In the proposed method, learning is performed so that only areas necessary for human recognition are observed.
% Therefore, it seems that the best evaluation value could not be obtained with the deletion score in ResNet18.

%表~\ref{tab:cub_del_in}に従来の視覚的説明の手法と提案手法のDeletionスコアとInsertionスコアの比較を示す．
%CUB-200-2010 datasetにおけるDeletionスコアとInsertionスコアの評価では，testサンプルを用いて評価する．
%表~\ref{tab:cub_del_in}から提案手法がABNよりも良い評価値を獲得していることが確認できる．
%したがって，従来のABNよりも説明性の高いAttention mapが獲得できていることがわかる．
%また，Grad-CAMと提案手法を比較するとResNet34とResNet50では良い評価値を獲得していることが確認できる．
%しかし，ResNet18におけるDeletionスコアではGrad-CAMが良い評価値であった．
%Grad-CAMは，認識対象の物体全体を注視する傾向が強い．
%一方で，ABNの注視領域はGrad-CAMよりも局所的である．
%また，提案手法では人が認識する上で必要な領域のみを注視するように学習させている．
%そのため，ResNet18におけるDeletionスコアでは最も良い評価値を獲得できなかったと考えられる．

In Table~\ref{tab:cub_coincidence}, we compare the degree of similarity between the attention maps output by the conventional visual explanation method and the proposed method and the attention maps created by bubble information. As shown in the table, the proposed method outputs the attention map that is closest to the one created by the bubble information. These results demonstrate that the proposed method can successfully embed human knowledge and output an attention map that contains this knowledge.

%表~\ref{tab:cub_coincidence}に従来の視覚的説明の手法と提案手法が出力したAttention mapとbubble情報により作成したAttention mapとの類似度の比較を示す．
%Attention mapの類似度による評価では，CUB-200-2010 datasetの学習サンプルを用いて評価する．
%表~\ref{tab:cub_coincidence}より提案手法が最もbubble情報により作成したAttention mapに近いAttention mapを出力していることが確認できる．
%したがって，提案手法により人の知見を考慮したAttention mapが出力できていることがわかる．

\begin{figure}[tb]
\centering
\includegraphics[width=0.95\linewidth]{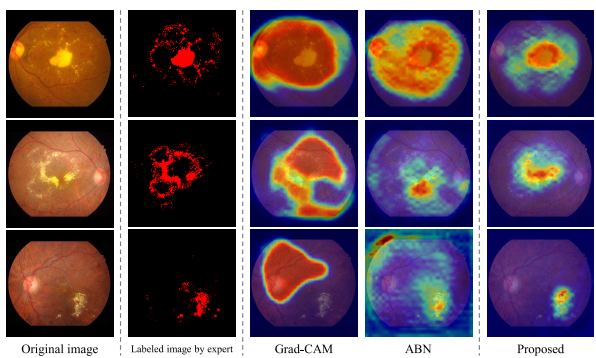}
\caption{Examples of conventional and proposed attention maps on IDRiD.}
\label{fig:retina_attention_map}
\end{figure}

\subsection{Fundus image classification on IDRiD}
Table~\ref{tab:comp_retina_recog} lists the classification accuracies on IDRiD. As shown, the ABN-based networks (e.g., AlexNet+ABN, ResNet$\ast$+ABN) achieved higher classification performances than the original networks. Moreover, by introducing the proposed fine-tuning method, the classification performances are further improved.
%
% 表~\ref{tab:comp_retina_recog}に，AlexNet，AlexNet+ABN，提案手法の疾患識別における精度を示す．
% 表~\ref{tab:comp_retina_recog}の結果より，従来のネットワーク（AlexNet, ResNet18, and ResNet34）に比べて，ABNを導入することで認識性能が向上している．
% さらに，提案するfine-tuningを適用することで，さらに精度が向上していることが確認できる．

Figure~\ref{fig:retina_attention_map} shows examples of the resultant attention maps. In the case of the conventional Grad-CAM, the attention maps broadly highlight both disease and non-disease regions. Also, the conventional ABN focuses on the non-disease regions around the disease regions. In contrast, the proposed method suppresses the highlighting for non-disease regions and focuses only on disease regions.
%
% 図~\ref{fig:retina_attention_map}に，AlexNet+ABNと提案手法のAttention mapを示す．
% AlexNet+ABNでは，疾患領域を含んだ非疾患領域に対してもAttention mapが強く反応している．
% IDRiDの学習サンプル数は深層学習を学習するには非常少ないため過学習が起きやすい.
% また,疾患領域周辺の非疾患領域に対してもAttention mapが反応していることがわかる．
% 一方，提案手法では非疾患領域の注視領域が減少し，疾患領域のみに注視領域が強く反応している．

We evaluate the obtained attention maps quantitatively. Tables~\ref{tab:idrid_del_in} and \ref{tab:idrid_coincidence} show that deletion and insertion scores and the similarity of the attention maps, respectively.
As shown in the Table~\ref{tab:idrid_del_in}, the proposed method has higher insertion scores than the other methods. As shown in the Table~\ref{tab:idrid_coincidence}, the proposed method outputs the attention map that is closest to the one edited by a segmentation label of disease regions annotated by a specialist. These results demonstrate that the proposed method is effective for fundus image recognition, where it is difficult to collect a large amount of training data, and that the interpretability of the attention map can be improved.
%
% これらの結果から，提案する深層学習ベースのHITLは医用画像認識のように大量のデータを準備するのが困難な画像認識タスクにおいても有効であり，視覚的説明に用いるAttention mapの改善が可能である．

% unused
% そのため，医用画像認識のように大量のデータを準備するのが困難な画像認識タスクでも，提案手法が有効であることがわかる．
% そのため，少ないデータからの学習では，注視領域が疾患部位からかけ離れる，または診断に無関係な領域を多く含み，自動診断の機能として信頼性が大きく損なわれることになる．

% \section{Discussion}
% \subsection{Improvement of our human in the loop}
% CUB-200-2010 datasetではtop-1 accuracyが最大でxxx~$\%$の向上が確認できたが，ImageNet datasetとIDRiDでは大きな改善が見られなかった．
% ImageNet datasetの場合，誤認識したサンプルのほとんどが「1つの画像に複数物体が映る」「ラベルミス」であるため，精度改善が少ないと考えられる．
% ラベルミスは，人手により最適なAttention mapを獲得しても，正しい認識結果を得ることができない．
% また，複数物体が映っている画像は注視領域を変更した場合でも，異なる物体に注視することで簡単に誤認識してしまう．
% そのため，本アプローチを適応しても，CUB-200-2010 Datasetのような大幅な精度向上が確認できなかったと考えられる．

% IDRiDでは，医師等の専門家でも判断が困難な疾患画像が多く含まれるため，専門家によるAttention mapの修正においても大幅な精度向上が困難である．
% しかし，患者によっては特殊な症例もあり，これらの症例に対してはFig.~\ref{fig:retina_attention_map}のように提案手法により疾患領域のみに注視領域が反応した信頼性の高いAttention mapと認識結果を得ることが可能である．

%また，本論文のアプローチでは，深層学習の注視領域と人の注視領域を同一にすることである．
%これにより，例外的に誤認識されたサンプルを人(専門家)の知見により矯正し，柔軟に例外的なサンプルをネットワークに学習させることができる．

% \subsection{Expert level of subject}
% ImageNet datasetにおいては開発したツールによりAttention mapによる認識精度を確認しつつ，インタラクティブに修正することで，品質を担保している．
% CUB-200-2010 datasetでは，人の知見により付与されたbubble情報を用いてAttention mapを作成している．この情報は，実際に人が複数種の鳥を認識するために注視した情報であり，高精度な認識結果を得るために必要な人の知見である．
% このbubble情報を用いて作成したAttention mapは人の注視した領域と同一であり，Attention mapの品質は高いと考える．
% 医用画像認識では，その専門家によりAttention mapを修正してもらうことを前提としているためAttention mapの専門性は保証できる．

\section{Conclusion}
In this paper, we proposed an approach to embed human knowledge into deep learning models by fine-tuning the network with a manually edited attention map.
%The proposed method takes advantage of a characteristic of ABN that enables adjustment of the recognition result corresponding to the manual edit of an attention map.
Specifically, the proposed method fine-tunes the ABN by calculating the training loss between the output attention map and the edited attention map. By fine-tuning using a manually edited attention map
by a human expert, we can embed human knowledge into the network and obtain an appropriate attention map for better visual explanation. Moreover, by introducing human knowledge to the attention map, classification performance is improved. 
% todo: ここを削るかも
Experimental results showed that the top-1 error on the ImageNet dataset was improved by 0.28$\%$, classification accuracy on the CUB-200-2010 dataset by 7.16$\%$, and classification accuracy on IDRiD by 3.67$\%$. 
%And, classification accuracies on CUB-200-2010 dataset and IDRiD were improved by xxx~$\%$ and xxx~$\%$, respectively.
%
% 本稿では，ABNが出力したAttention mapを用いた深層学習ベースのHITLを提案した．
% 提案手法では，ABNのAttention mapを人の知見により修正した場合，修正に応じて認識結果を調整できる特性に着目している．
% 提案する深層学習ベースのHITLでは，ABNが出力したAttention mapと修正したAttention mapから学習誤差を算出することで，ABNのブランチをファインチューニングする．
% これにより，提案手法は,人の知見を取り入れたAttention mapを出力することが可能となった.
% さらに,改善したAttention mapをAttention機構へ応用することで認識性能の向上を確認した．
% ImageNet datasetにおける一般画像認識ではxx~$\%$，CUB-200-2010 datasetにおける詳細認識では約xx~$\%$向上した.
%
Our evaluation of the attention maps showed that the proposed method obtained better deletion and insertion scores than conventional methods. Moreover, the similarity score results show that the proposed method can provides attention maps that are similar to the edited by a human expert. Consequently, our method can generate a more interpretable attention map and successfully embed human knowledge. 
% Attention mapの数値的な評価では，DeletionスコアとInsertionスコアにおいてABNよりも高い評価値を獲得した．
% また，提案手法により得られたAttention mapは従来の視覚的説明と比較して最も人の注視領域との類似性が高いことが確認できた．
% 提案手法によって得られるAttention mapは人の知見を取り組んでおり，視覚的説明に適している.
%
%今後は，HITLのさらなる高精度化に取り組む．（藤吉先生の相談）
Our future work will include further improvement of the performance by editing attention maps with multi-resolution.

{\small
\bibliographystyle{ieee_fullname}
\bibliography{root}
}

\end{document}